\documentclass[letterpaper]{article} % DO NOT CHANGE THIS
\usepackage{aaai2026}  % DO NOT CHANGE THIS
\usepackage{times}  % DO NOT CHANGE THIS
\usepackage{helvet}  % DO NOT CHANGE THIS
\usepackage{courier}  % DO NOT CHANGE THIS
\usepackage[hyphens]{url}  % DO NOT CHANGE THIS
\usepackage{graphicx} % DO NOT CHANGE THIS
\usepackage{natbib}  % DO NOT CHANGE THIS AND DO NOT ADD ANY OPTIONS TO IT
\usepackage{caption} % DO NOT CHANGE THIS AND DO NOT ADD ANY OPTIONS TO IT
\usepackage{algorithm}
\usepackage{algorithmic}

\usepackage{newfloat}
\usepackage{listings}
\DeclareCaptionStyle{ruled}{labelfont=normalfont,labelsep=colon,strut=off} % DO NOT CHANGE THIS
\floatstyle{ruled}
\newfloat{listing}{tb}{lst}{}
\floatname{listing}{Listing}
\usepackage[colorlinks=true]{hyperref} % -- This package is specifically forbidden

\usepackage{makecell,pifont,multirow}
\usepackage{colortbl}
\usepackage{caption}
\usepackage{booktabs}
\usepackage{subcaption}
\usepackage{amssymb}
\usepackage{marvosym}
\usepackage{titletoc}
\usepackage{amsmath}

\def\netName{LMAD}

\title{{\em \netName{}}: Integrated End-to-End Vision-Language Model \\ for Explainable Autonomous Driving}

\author {
    Nan Song\textsuperscript{\rm 1}\quad
    Bozhou Zhang\textsuperscript{\rm 1}\quad
    Xiatian Zhu\textsuperscript{\rm 2}\quad
    Jiankang Deng\textsuperscript{\rm 3}\quad
    Li Zhang\textsuperscript{\rm 1}\corrauthor
}
\affiliations {
    \textsuperscript{\rm 1}School of Data Science, Fudan University\quad
    \textsuperscript{\rm 2}University of Surrey\quad
    \textsuperscript{\rm 3}Imperial College London
    \\
    \vspace{.4em} 
}

\usepackage{bibentry}
\begin{document}

\maketitle

%%%%%%%%%%%%%%%%%%%%%%%%%%%%%%%%%%%%%%%%%%%%%%%%%
 \begin{abstract}
Large vision-language models (VLMs) have shown promising capabilities in scene understanding, enhancing the explainability of driving behaviors and interactivity with users. Existing methods primarily fine-tune VLMs on on-board multi-view images and scene reasoning text, but this approach often lacks the holistic and nuanced scene recognition and powerful spatial awareness required for autonomous driving, especially in complex situations. To address this gap, we propose a novel vision-language framework tailored for autonomous driving, called \textbf{\netName{}}. Our framework emulates modern end-to-end driving paradigms by incorporating comprehensive scene understanding and a task-specialized structure with VLMs. In particular, we introduce preliminary scene interaction and specialized expert adapters within the same driving task structure, which better align VLMs with autonomous driving scenarios. Furthermore, our approach is designed to be fully compatible with existing VLMs while seamlessly integrating with planning-oriented driving systems. Extensive experiments on the DriveLM and nuScenes-QA datasets demonstrate that \netName{} significantly boosts the performance of existing VLMs on driving reasoning tasks, setting a new standard in explainable autonomous driving.

\end{abstract}    
\section{Introduction}
\label{sec:intro}

The rapid development of autonomous driving has stimulated the research across various domains, with particular emphasis on enhancing scene understanding and behavior explainability. Recently, large vision-language models (VLMs)~\cite{liu2023llava, zhang2023llamaadapter, gao2023llamaadapterv2, chen2024internvl} have emerged as powerful tools in bridging visual and linguistic information, showcasing considerable potential in explaining driving behaviors and improving interactions with users. Hence, the research on incorporating VLMs into autonomous driving is undergoing extensive exploration. Recent methods directly fine-tune VLMs with proposed datasets~\cite{wang2023drivemlm, nie2023reason2drive, sima2023drivelm, tian2024drivevlm, huang2024drivemm, xie2025drivebench} or employ powerful and off-the-shelf language models to construct agents~\cite{wen2023dilu, mao2023gptdriver, ma2023dolphins}, facilitating the steady progress of explainable autonomous driving.

\begin{figure}[ht]
    \centering
    \includegraphics[width=0.46\textwidth]{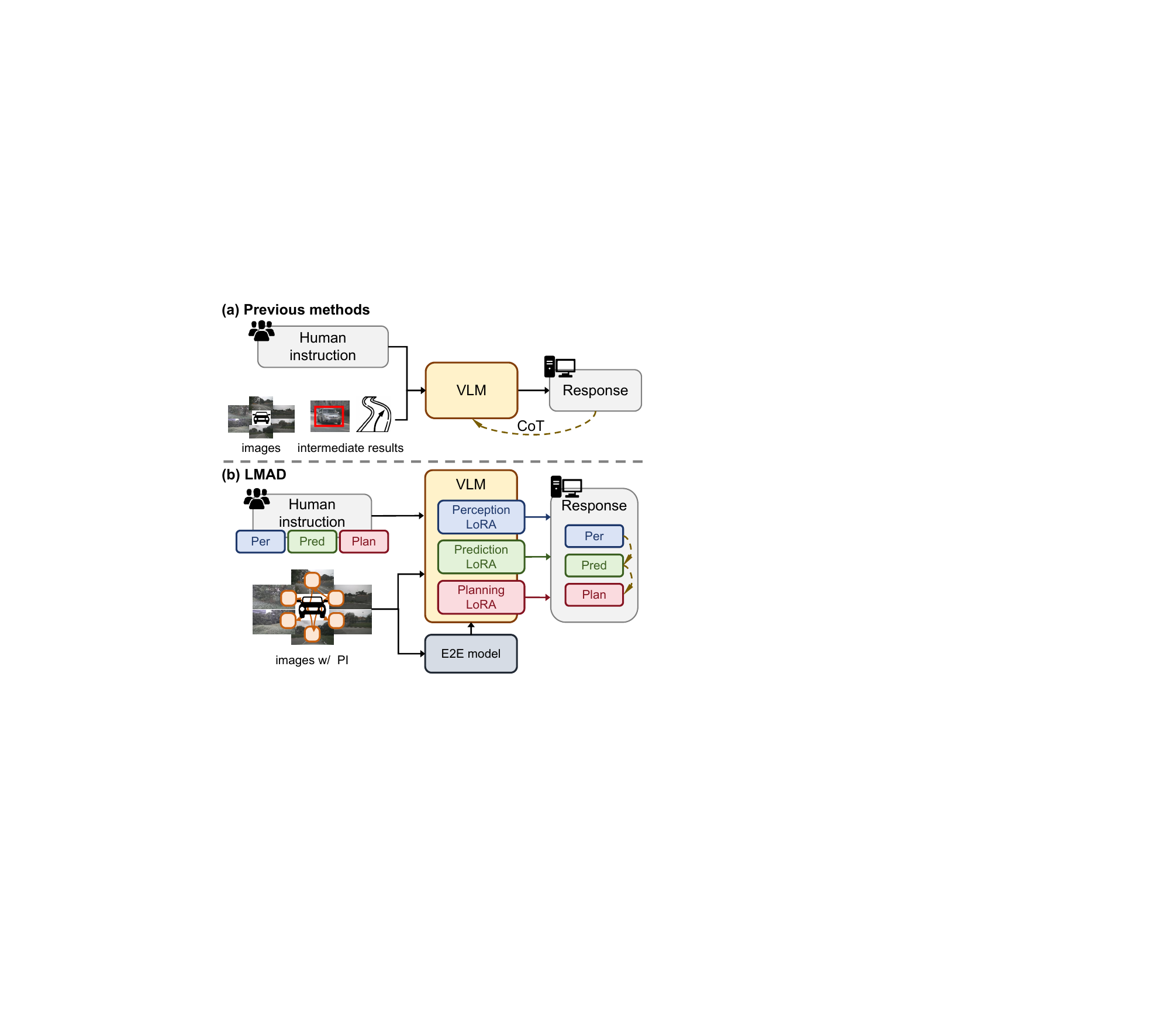}
    \caption{Comparison of (a) previous methods that employ standard VLMs on driving scenarios and (b) our LMAD with a task-specialized structure and the scene preliminary interaction mechanism.}
    \label{fig:compare}
    % \vspace{-3mm}
\end{figure}

% Despite these advances, they still face challenges in end-to-end behavior reasoning considering the intrinsics of autonomous driving, i.e., holistic surrounding scenes and progressive driving tasks. 

Drawing inspirations from autonomous driving frameworks, recent approaches use simple visual representation and intermediate results of driving system to provide scene information, and directly employ or finetune VLMs with optional Chain-of-Though (CoT) for reasoning~\cite{sima2023drivelm, ma2023dolphins, sha2023languagempc, qian2025agentthink}, as shown in Figure~\ref{fig:compare} (a). However, directly using this information makes it challenging for the model to capture relationships between traffic elements, thereby impeding a holistic understanding of driving scenes. In addition, the VLM still struggles in localization and motion estimation, which are crucial for autonomous driving behavior analysis. During progressive reasoning process, this might cause severe cumulative errors, thus leading to suboptimal performance of existing methods in driving tasks.

% Despite these advances, they still face challenges in end-to-end behavior reasoning considering the intrinsics of autonomous driving, i.e., holistic surrounding scenes and progressive driving tasks. To resolve these issues, recent approaches draw inspirations from autonomous driving framework, introducing BEV to provide detailed scene representations~\cite{qian2024nuscenes, shao2024lmdrive} and employing Chain-of-Thought (CoT) and relevant variants to perform step-by-step reasoning~\cite{sima2023drivelm, ma2023dolphins, sha2023languagempc} as shown in~\cref{fig:compare} (a). However, on one hand, the BEV representations might conflict with human-perspective reasoning and ignore with critical influence on the scene, such as traffic signs. In addition, VLMs struggle in localization and motion estimation tasks, which are crucial for comprehensive scene understanding in autonomous driving. Hence, existing methods tend to cause confusion and cumulative errors, exhibiting suboptimal performance in reasoning.

Addressing these limitations, we propose a novel framework \netName{} tailored for autonomous driving scenarios (Figure~\ref{fig:compare} (b)), leveraging both holistic scene understanding and optimized task-specialized structure of modern end-to-end driving paradigms. 
We introduce a Preliminary Interaction (PI) mechanism for prior scene relationship modeling and integrate a VLM with a set of specialized expert adapters (e.g., LoRA \cite{hu2022lora})
in the same driving task structure encompassing perception, prediction and planning in sequence. The PI mechanism can provide rough relational information of traffic participants and alleviate the learning complexity of VLMs, while adapters work collaboratively to address questions of varying complexity, which is achieved by their targeted acquisition of task-specific driving knowledge from driving scenarios. In addition, \netName{} also facilitates seamless incorporation of prior knowledge from an end-to-end driving system,
further complementing with VLMs and enhancing subsequent reasoning.
% We first enhance VLMs with enriched scene information and the specialization in each driving tasks through Multi-view QFormer and Parallel LoRA fine-tuning. 
% Specifically, 
% the QFormer~\cite{li2023blip2} is widely utilized in driving VLMs~\cite{nie2023reason2drive, wang2023drivemlm, shao2024lmdrive} as vision feature compression module. 
% To retain the critical information of driving scenes and provide prior contextual corelation during compression, we improve QFormer by adjusting the interactive relationships and enable the learnable queries to observe entire scenes. 
% For the fine-tuning process, we present Parallel LoRA for handling different type tasks, which facilitates the task-specific proficiency by introducing only a few additional parameters. Benefiting from these modules, VLMs can comprehensively understand the scenes and perform progressive reasoning like an end-to-end driving framework. Moreover, to better align VLMs with end-to-end models, we also establish connections between them through the interaction of intermediate features. This provide rich driving prior information and complement each other with Parallel LoRA, effectively boosting the spatial awareness for VLMs.

Our \textbf{contributions} are summarized as follows:
\textbf{(i)} We propose leveraging the advanced design philosophy of end-to-end driving paradigms
to achieve explainability in autonomous driving.
% Motivated by end-to-end planning frameworks in autonomous driving, We introduce \netName{}, a novel integrated VLM framework equipped with Multi-view QFormer and fine-tuned with Parallel LoRA, advancing the holistic scene understanding and the divergence of different driving task. 
\textbf{(ii)}
We introduce \netName{}, a novel VLM framework characterized with the scene preliminary interaction mechanism and a task-specific expert structure,
allowing seamless incorporation of prior knowledge from an existing end-to-end driving system.
% equipped with Multi-view QFormer and fine-tuned with Parallel LoRA, advancing the holistic scene understanding and the divergence of different driving task. 
% We introduce integrate end-to-end planning frameworks with \netName{}, providing driving knowledge for VLMs and compensating for the shortcomings in localization and motion estimation.
\textbf{(iii)}
Extensive experiments on the DriveLM and nuScenes-QA reasoning benchmarks demonstrate that \netName{} achieves state-of-the-art performance.

\section{Related work}
\label{sec:related_work}

\paragraph{VLMs for autonomous driving.}
With the emergence of GPT series~\cite{openai2023gpt}, LLMs have been successfully applied across various domains, yielding significant improvements due to their powerful reasoning abilities and extensive world knowledge. Recently, the field has witnessed an increased focus on VLMs, which incorporate vision encoders~\cite{li2023blip2} and further unleash the capabilities of LLMs. Some approaches~\cite{zhu2023minigpt, liu2023llava, bai2023qwen,chen2024internvl} leverage more advanced architectures and richer datasets to achieve superior multi-modal performance, though they often neglect the associated resource and time costs. In contrast, to strike a balance between performance and efficiency, Parameter-Efficient Fine-Tuning (PEFT) methods~\cite{zhang2023llamaadapter, gao2023llamaadapterv2, alayrac2022flamingo} tackle this challenge by introducing a limited number of learnable parameters, such as Prompt-Tuning, Adapter, and LoRA~\cite{hu2022lora}.

In the field of autonomous driving, VLMs can function as human experts, promoting the explainability of driving behaviors and addressing challenging long-tail and zero-shot scenarios. Recently, tens of thousands of driving scenarios with language annotations have been made available~\cite{chen2023driving, nie2023reason2drive, qian2024nuscenes, sima2023drivelm, xie2025drivebench}, enabling the exploration of VLMs in autonomous driving. Some methods~\cite{mao2023gptdriver, wen2023dilu, wang2024omnidrive, xing2025openemma} directly resort to robust GPT series to construct driver agents capable of performing driving tasks, effective but constrained by the limitations of GPT models. To overcome this challenge, latest work~\cite{shao2024lmdrive, xu2023drivegpt4, wang2023drivemlm} has focused on training specialized VLM models, ensuring the scalability and generalizability of these frameworks. Especially, DriveLM~\cite{sima2023drivelm} and its follow-up work~\cite{xie2025drivebench} build a distinct Graph-of-Thoughts technique that spans from perception to planning, thus improving comprehensive reasoning capabilities across various tasks in autonomous driving. Recently, DriveMM~\cite{huang2024drivemm} aggregates various driving language datasets to create a general VLM for autonomous driving. However, these fine-tuned VLMs still lack the task-specific knowledge or spatial awareness required to tackle complex scene reasoning.

Additionally, several works have sought to bridge classical end-to-end frameworks~\cite{jiang2023vad, hu2023planning} and language models. VLP~\cite{pan2024vlp} and VLM-AD~\cite{xu2024vlm} supervise end-to-end framework with auxiliary contrastive learning between the intermediate features of the driving output and text or the planning results of driving model and VLMs. Moreover, DriveVLM~\cite{tian2024drivevlm}, Senna~\cite{jiang2024senna} and Hint-AD~\cite{ding2024hintad} deliver some specific outputs or features between these two components to support explainable reasoning. Besides being integrated with end-to-end models, our method, in contrast, our framework enhances VLMs by promoting scene interaction and fostering specialization in each driving task, which further improves the alignment between VLMs and end-to-end autonomous driving frameworks.

\begin{figure*}
    \centering
    \includegraphics[width=0.98\textwidth]{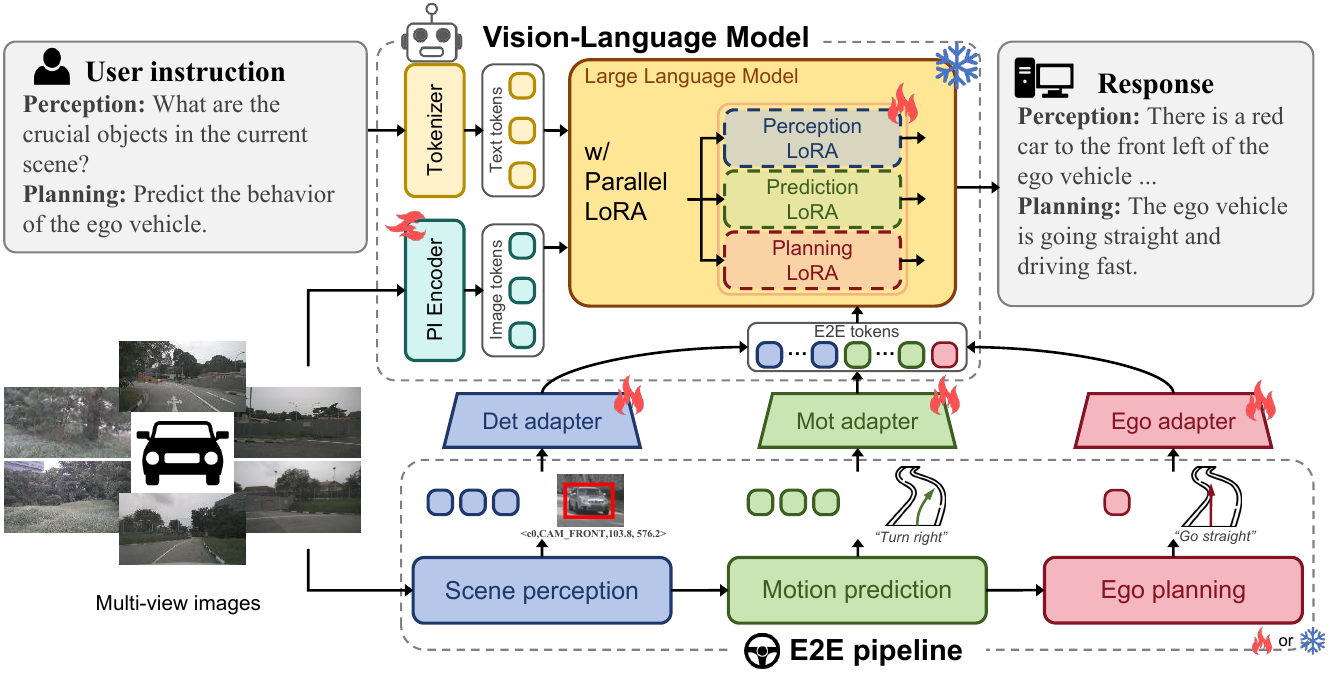}
    \caption{\textbf{Schematic illustration of \netName{}}, which integrates an end-to-end driving pipeline with the vision-language model. The text and image tokens are extracted from tokenizer and then interacted by PI encoder. The end-to-end tokens derive from intermediate features and corresponding prompts, subsequently processed by respective adapters and aggregated. The language model is fine-tuned with Parallel LoRA with all these tokens, better aligned with end-to-end autonomous driving.}
    \label{fig:pipeline}
\end{figure*}
\paragraph{End-to-end autonomous driving.}
Rather than focusing on the individual tasks in the field of autonomous driving, recent approaches~\cite{casas2021mp3, hu2022stp3, hu2023planning, jiang2023vad} are delving into the end-to-end planing that can simultaneously handling joint tasks, from scene perception to ego-planning, within a unified framework. Earlier work~\cite{casas2021mp3, hu2022stp3} achieves this framework with simplified intermediate tasks, which limits the overall planning performance. UniAD~\cite{hu2023planning} makes significant strides in end-to-end planning by covering a broad spectrum of driving tasks, followed by VAD~\cite{jiang2023vad}, which enhances the framework with vectorized scene representations and optimized module composition. Benefiting from explicit and comprehensive intermediate results, these methods realize a remarkable breakthrough in the planning task. Besides, some latest approaches~\cite{sparsead, sparsedrive} introduce the sparse representation to efficiently perform end-to-end planning. In this work, we aim to reconstruct VLMs by imitating existing end-to-end driving models, utilizing preliminary interaction and task-specialized reasoning for improved performance. Besides, we also explore the benefits that VLMs offer to end-to-end models.

\section{Methodology}

In this section, we present \netName{}, an integrated vision language model with end-to-end planning frameworks for explainable autonomous driving as illustrated in Figure~\ref{fig:pipeline}.

\subsection{End-to-end vision language model}
\label{sec:vlm}

We adopt a universal VLM framework as our foundational model, consisting of a vision encoder to extract visual tokens, a tokenizer to encode text tokens, and a language decoder to generate responses based on these visual and textual inputs. To effectively address the extensive and progressive end-to-end QA tasks~\cite{sima2023drivelm, nie2023reason2drive}, we further adapt VLMs for end-to-end autonomous driving models, which take as input multi-view surrounding images and handle a wide range of driving tasks. To achieve this, we present the scene Preliminary Interaction (PI) encoder and the Parallel LoRA module to efficiently aggregate scene relationships of traffic elements and integrate task-specific driving knowledge, respectively.

\paragraph{Preliminary interaction encoder.}
In autonomous driving scenarios, multi-view images are frequently employed to facilitate comprehensive attention to the entire scene. However, independently processing each image, as commonly done in VLMs, generates a substantial number of less relevant cross-view image tokens, increasing the learning burden due to complex spatial relationships. To mitigate this issue, BEV feature maps have been adopted as a holistic representation in recent driving-focused VLMs~\cite{qian2024nuscenes, shao2024lmdrive}. Although effective, it heavily relies on the specialized pretrained BEV encoder and often overlooks detailed traffic information (such as traffic signs), ultimately constraining performance in scene comprehension tasks.

\begin{figure}
    \centering
    \includegraphics[width=0.47\textwidth]{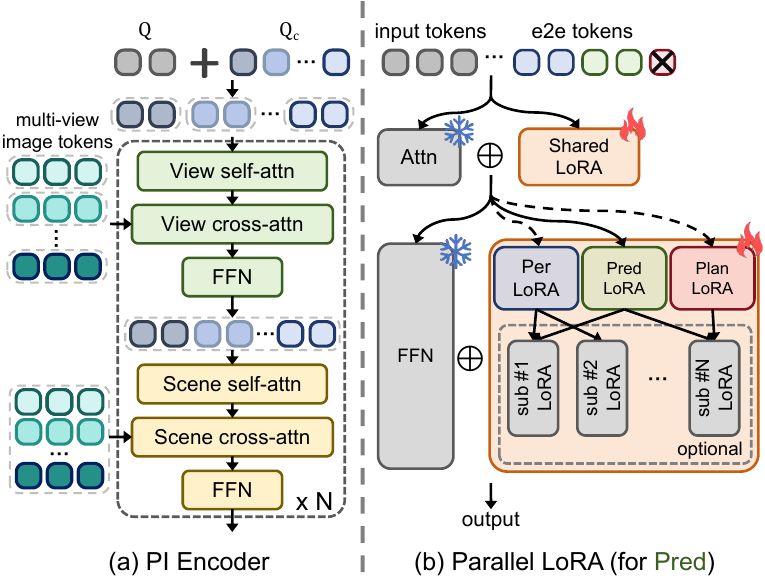}
    \caption{\textbf{Schematic illustration} of (a) PI encoder and (b) Parallel LoRA.
    The dashed input into Parallel LoRA represents inactive for current question.}
    \label{fig:module}
\end{figure}

Hence, we retain the multi-view input and propose the PI encoder to model preliminary scene relationships, which introduces decoupled queries and alternating attention mechanisms as illustrated in Figure~\ref{fig:module} (a). Specifically, we employ decoupled queries to collect scene information, consisting of $N_{q}$ general vision queries $Q$ and $N_{c}$ camera queries $Q_{c}$. The vision queries $Q$ are utilized to capture image context, while camera queries $Q_{c}$ serve as identifiers for distinct camera perspectives, which assist in addressing perspective-specific questions and constructing spatial relationships. Subsequently, vision queries and camera queries are integrated through broadcast addition to form the overall $N_{q}\times N_{c}$ queries, and then propagated to alternating attention blocks to facilitate both view-level and scene-level information interactions. In the odd-numbered blocks, we partition all queries into $N_{c}$ specific groups for each camera, and allow interactions only within each group and with their corresponding image features, thereby fully preserving the unique information of each image. In the even-numbered blocks, all queries are combined together to perform scene-level self-attention with each other and cross-attention with all multi-view image features. Notably, we apply slightly different strategies for the Qformer-based models~\cite{gao2023llamaadapterv2} and direct input models~\cite{bai2023qwen, chen2024internvl}. For the former, we introduce additional learnable camera queries and modify the interactive relationships, ensuring compatibility with powerful pretrained weights. In contrast, for direct input methods that originally lack an extra vision compression encoder, we replace camera queries $Q$ with multi-view image tokens and use lightweight alternating self-attention blocks to achieve the interaction.

\paragraph{Fine-tuning with Parallel LoRA.}
To endow VLMs with the reasoning abilities in the field of autonomous driving, we also employ LoRA~\cite{hu2022lora} modules for efficient fine-tuning, which can preserve the generalization and world knowledge of pretrained VLMs. In the end-to-end settings, models must tackle a variety of planning-related driving tasks, each focusing on different aspects of the driving process. Hence, we propose the Parallel LoRA (P-LoRA) module to handle diverse tasks and improve task proficiency,  imitating the existing multitask driving paradigm.

As shown in Figure~\ref{fig:module} (b), the P-LoRA module is implemented by replacing the vanilla LoRA module in FFN blocks with several parallel LoRA branches, each responsible for a task to collect task-specific knowledge. For the LoRA equipped in attention blocks, however, we keep them unchanged and shared, preserving general driving knowledge. In our implementation, we set up individual LoRA branches for perception, prediction, and planning, respectively, activating the corresponding branch and end-to-end tokens based on the task type of input questions. For instance, the decoder can only access detection and prediction tokens to address prediction tasks. This design allows a single language decoder to be transformed into multiple specialized task heads, while adding minimal trainable parameters. In addition, we offer an optional hierarchical format designed to adapt to various types of both driving tasks and QA tasks, providing specialization for datasets that involve complicated instructions. In the inference stage, we incorporate the CoT reasoning technique into the P-LoRA module, following end-to-end methods to output results step by step.

\subsection{Integrated with end-to-end driving}
\label{sec:with_e2e}
The features of end-to-end autonomous driving frameworks can provide rich object position and motion priors for VLMs, enhancing the spatial awareness and reasoning capabilities. To fulfill this, we collect the output features of perception, prediction and planning as feature tokens $F_{\rm det}$, $F_{\rm mot}$ and $F_{\rm ego}$, respectively. Concretely, the top $N_{\rm ins}$ objects are selected to yield the $F_{\rm det}$ and $F_{\rm mot}$ based on detection confidence, while the unique token $F_{\rm mot}$ for planning is adopted. Furthermore, we also provide numerical and text prompts for each token to facilitate the model understanding. For the numerical prompts of prediction and planning, we employ Multi-Layer Perceptron (MLP) to project predicted trajectories $T_{\rm mot}$ of scene objects and ego planned trajectory $T_{\rm ego}$ to high-dimensional features $F_{\rm mot}^{\rm n}$ and $F_{\rm ego}^{\rm n}$. In addition, the corresponding text prompts are derived according to the steering and speed variations, forming the text such as ``\textit{Go straight, acceleration}'', ``\textit{Turn Right, constant speed}'' and so on. And then, text prompts are fed into a simple Multi-Head Attention module with an additional query to generate text prompt features. This process can be formulated as:
\begin{equation}
F_{(\ast)}^{\rm n} = {\rm MLP}(T_{(\ast)}), \; F_{(\ast)}^{\rm t} = {\rm MHA}(F_{q}, \Tilde{F}_{(\ast)}^{t}),
\end{equation}
where $\ast$ refers to $mot$ or $ego$, $F_{(\ast)}^{\rm t}$ and $\Tilde{F}_{(\ast)}^{t}$ refer to text prompt features and original text features encoded by input embedding module of language models, and $F_{q}$ is the learnable query to aggregate text information.

As for the detection prompts, we project the detected 3D bounding boxes onto 2D images, yielding normalized 2D boxes and depth. The numerical prompt features $F_{\rm det}^{\rm n}$ are directly calculated from these properties with a MLP module, while we construct text prompts based on the defined format of different datasets and process using the same method as above. After obtaining all prompt features, we incorporate them into corresponding tokens through add operation. Subsequently, we employ several adapters $\mathcal{A}$ to independently process these three kinds of token features and align them with language context, finally outputting the concatenation of features as overall end-to-end tokens $F_{e2e} \in \mathbb{R}^{(2N_{\rm ins} + 1) \times D}$:
\begin{equation}
F_{e2e} = \mathcal{A}_{\rm det}(F_{\rm det})\oplus\mathcal{A}_{\rm mot}(F_{\rm mot})\oplus\mathcal{A}_{\rm ego}(F_{\rm ego}),
\end{equation}
where $D$ is the feature dimension of language models. We then append the end-to-end tokens following the vision tokens and pass them into the language decoder. It is worth noting that we adopt different implementation methods for this feature transmission, which depend on the specific language models. In detail, end-to-end tokens are treated as adapters, similar to vision tokens in the LLaMA-Adapter model~\cite{gao2023llamaadapterv2}, while serving as input tokens in other mainstream models such as LLaVA~\cite{liu2023llava} and InternVL~\cite{chen2024internvl}.

\subsection{Model training}
\label{sec:train}
In our framework, we employ two training strategies for the end-to-end driving and language branches. For the first one, the overall end-to-end driving branch is frozen while the language branch is fine-tuned, by which we anticipate to enhance the reasoning and understanding abilities of language model in the field of autonomous driving. In this case, the end-to-end tokens are detached, and the model is trained solely with text generation supervision, where auto-regressive cross-entropy loss $\mathcal{L}_{\rm txt}$ is used.

\begin{table*} [ht!] 
    \centering
        \centering
        % \resizebox{0.98\columnwidth}{!}
        {\begin{tabular}[b]{l|c|cccccc|c}
        \toprule[1.5pt]

        \multirow{2}{*}{\textbf{Method}} & \multirow{2}{*}{\textbf{Size}}& \multirow{2}{*}{Acc.$\uparrow$} & \multirow{2}{*}{GPT $\uparrow$} &
        \multicolumn{3}{c}{Language Score $\uparrow$} & \multirow{2}{*}{Match $\uparrow$} & \multirow{2}{*}{Final $\uparrow$} \\
         & & & & $\rm BLEU$ & $\rm ROUGE\_L$ & $\rm CIDEr$ & & \\
        \midrule 

        LLaMA-Adapter~\cite{gao2023llamaadapterv2}&
        \multirow{2}{*}{7B} &
        70.14 & 53.73 & 65.20 & 72.84 & 0.86 & 34.19 & 51.62 \\
        % \rowcolor{gray!10}
        $+$ LMAD& &
        \bf 72.55& \bf 55.82& 65.03 & 72.73 & \bf 1.03 & \bf 35.48 & \bf 53.19 \\
        \hline
        LLaVA-1.5~\cite{liu2023llava}&
        \multirow{2}{*}{7B} &
        72.39& 61.91 & 65.32 & 73.91 & 0.97 & 34.41 &  55.47\\
        % \rowcolor{gray!10}
        $+$ LMAD& &
        \bf74.56& \bf63.80& \bf 67.87 & \bf74.14 & \bf1.70 & \bf35.19 & \bf57.05 \\
        \hline
        InternVL2~\cite{chen2024internvl}&
        \multirow{2}{*}{4B} &
        77.95 & 64.13 & 68.21 & 73.57 & 1.83 & 42.52 & 59.31 \\
        % \rowcolor{gray!10}
        $+$ LMAD& &
        \bf 80.38 & \bf 65.10 & \bf 69.51 & \bf 74.10 &  1.78 & \bf 46.12 & \bf 61.03 \\

         \bottomrule[1.5pt]
        \end{tabular}}
    
    % \vspace{1.5pt}
    \caption{Performance on \textit{DriveLM official benchmark}. For each metric, the improved results are in \textbf{bold}. \netName{} are integrated with three baseline VLMs and can lead to consistent improvements.}
    \label{tab:drivelm}
\end{table*}
\begin{table} [ht!] 
    \centering
        \centering
        \resizebox{0.98\columnwidth}{!}
        {\begin{tabular}[b]{l|cccc}
        \toprule[1.5pt]

         Method & \rm BLEU4 & \rm ROUGE L & \rm CIDEr &\rm METEOR \\
        \midrule 

        DriveLM-Agent&53.09 & 66.79 & 2.79 & 36.19 \\
        EM-VLM4AD&45.36 & 71.98 & 3.20 & 34.49 \\
        MiniDrive&50.20 & 73.50 & 3.32 & 37.40 \\
        LLaMA-Adapter&45.96 & 69.78 & 3.07 & 33.66 \\
        MPDrive&52.71 & \bf 76.98 & 3.56 & 38.31 \\
        LMAD&\bf54.59 & 75.72 & \bf 3.84  & \bf 39.24 \\

         \bottomrule[1.5pt]
        \end{tabular}}
    
    % \vspace{1.5pt}
    \caption{Performance comparison on the DriveLM set following EM-VLM4AD.}
    \label{tab:drivelmcmp}
\end{table}
\begin{table} [t!] 
    \centering
    {\begin{tabular}{l|ccc}
       \bottomrule[1.5pt]
        $\textbf{Method}$ & Overall & H0 & H1\\\hline
        GPT-4o&
        37.1 & 42.0 & 34.7 \\ 
        Gemini 1.5&
        35.4 & 40.5 & 32.9\\
        ADAPT~\cite{jin2023adapt}&
        46.4 & 51.0 & 44.2\\
        MCAN~\cite{yu2019mcan}&
        49.9 & 56.2 & 46.7\\
        Vote2Cap~\cite{chen2024vote2cap}&
        47.0 &  51.2 & 44.9\\
        TOD$^{3}$Cap~\cite{jin2025tod3cap}&
        49.0 &  53.0 & 45.1\\
        Hint-AD~\cite{ding2024hintad}&
        50.5 &  55.4 & 48.0\\
        \Xhline{0.9pt} 
        \netName{}       & \bf 51.8 & \bf 56.5 & \bf 49.6 \\\bottomrule[1.5pt]
    \end{tabular}}
    % \vspace{1.5pt}
    \caption{Performance comparison on {\it nuScenes-QA test set}.}
    \label{tab:nuscqa}
\end{table}

Additionally, we aim to explore whether the language model can contribute to enhancing the end-to-end planning performance. However, VLMs typically generate high-level text commands in QA tasks, hardly providing precise trajectory information for end-to-end framework. To address this, we activate the gradient flow from the language branch to the end-to-end branch and train the decoder heads for the end-to-end task. This encourages the end-to-end framework to learn to focus more on the key elements that influence ego planning. The loss of this strategy can be formulated as:
\begin{equation}
\mathcal{L} = \mathcal{L}_{\rm txt} + \lambda\mathcal{L}_{\rm e2e},
\end{equation}
where $\lambda$ is the balancing factor and $\mathcal{L}_{\rm e2e}$ is composed of detection loss, motion prediction loss and planning loss aggregated with the default weights in the end-to-end model training settings. The experiments of this strategy are demonstrated in the supplementary material.

\section{Experiments}

\begin{table*} [ht!] 
    \centering
    \begin{tabular}{c|ccc|ccccc}
       \bottomrule[1.5pt]
         ID & PI encoder & P-LoRA & E2E tokens & Acc. & GPT Score & Lan. Score & Match & Final Score\\
         \hline
          
        1 &  &  &  & 71.88 & 62.23 & 45.85 & 34.66 & 55.37 \\ 
        2 & \checkmark &  &  & 72.51 & 62.94 & 46.69 & 34.47 & 55.91 \\
        3 & \checkmark & \checkmark &  & 73.39 & 63.40 & 47.56 & 34.92 & 56.53  \\
        4 & \checkmark &  & \checkmark & 74.18 & 63.03 & 47.12 & 35.09 & 56.49 \\
        \rowcolor{gray!20}
        5 & \checkmark & \checkmark & \checkmark & \bf 74.61 &\bf  63.96 &\bf  48.02 &\bf  35.28 &\bf  57.17 \\
        \bottomrule[1.5pt]
    \end{tabular}
    % \vspace{1.5pt}
    \caption{Ablation study on the core components of \netName{} on the {\it DriveLM validation set}. ``Lan. Score" indicates language score that is the average of BLEU, ROUGE\_L and CIDEr.}
        \label{tab:ab_main}
\end{table*}
\begin{table} [ht!] 
    \centering
    % \resizebox{\columnwidth}{!}{
    \begin{tabular}{c|c|cccc}
       \bottomrule[1.5pt]
         P-LoRA & Type & Acc. & GPT & Lan. & Match\\\hline
         &  & 72.51 & 62.94 & 46.69 & 34.47\\
        \checkmark & task & 73.39 & 63.40 & 47.56 & 34.90\\
        \checkmark & ques. & 73.52 & 63.51 & 46.48 &34.66\\
        \rowcolor{gray!20}
        \checkmark & hier. & \bf 73.95 & \bf  63.48 & \bf 47.83 & \bf 34.93\\
        \bottomrule[1.5pt]
    \end{tabular}
    % }
    % \vspace{1.5pt}
    \caption{Ablation study on the Parallel LoRA fine-tuning. ``ques.'' and ``hier.'' refer to the P-LoRA with question type (such as multiple-choice question and scene description question.) and hierarchical mode.}
    \label{tab:ab_lora}
\end{table}
\begin{table} [t!] 
    \centering
    % \resizebox{\columnwidth}{!}{
    \begin{tabular}{ccc|cccc}
       \bottomrule[1.5pt]
          Per. & Pred. & Plan & Acc. & GPT & Lan. & Match\\\hline
         & & & 71.88 & 62.23 & 45.85 & 34.66\\
         \checkmark & & & 73.07 & 62.59 & 46.54& \bf 34.87\\
         \checkmark & \checkmark & & 73.38 & 62.72 & 46.69 &34.83\\
        \rowcolor{gray!20}
         \checkmark & \checkmark & \checkmark & \bf 73.81 & \bf 62.93 & \bf 46.76& 34.85\\
        \bottomrule[1.5pt]
    \end{tabular}
    % }
    % \vspace{1.5pt}
    \caption{Ablation study on the end-to-end tokens. ``Per." and ``Pred." represent the perception and prediction tokens.}
        \label{tab:ab_e2etoken}
\end{table}

\subsection{Experimental settings}
\label{sec:expset}

\paragraph{Datasets and evaluation metrics.}
We assess the language reasoning performance of our method using the DriveLM~\cite{sima2023drivelm} and nuScenes-QA~\cite{qian2024nuscenes} datasets, all of which are VQA benchmarks labeled on the nuScenes~\cite{caesar2019nuscenes} autonomous driving dataset. The DriveLM dataset comprises 377,956 QA pairs of some key scenes selected from the entire datasets. For each scene, progressive QA pairs ranging from perception, prediction to planning and behavior are involved, providing comprehensive explanation of end-to-end driving behavior. The nuScenes-QA dataset encompasses approximately 460k QA pairs, primarily focusing on perception tasks. In addition to QA pairs, surrounding images with six perspectives are adopted same as the nuScenes. More details are provided in the supplementary materials.

\paragraph{Implementation details.}
Regarding the baseline, we integrate \netName{} with three types of VLMs, including LLaMA-Adapter-V2~\cite{gao2023llamaadapterv2}, LLaVA-v1.5~\cite{liu2023llava} and InternVL2~\cite{chen2024internvl}, to ensure versatility. For the end-to-end planning branch, we adopt VAD-base~\cite{jiang2023vad} framework, which is both reliable and relatively lightweight. Besides, for the experiments on nuScenes-QA dataset, we employ LLaMA-Adapter as the foundation following competitive methods. For model training, we leverage the AdamW~\cite{loshchilov2018decoupled} optimizer with a weight decay of 0.01. We utilize a cosine learning
rate decay scheduler with a warm-up ratio of 0.03. We fine-tune our model with a batch size of 16 on 8 A6000 GPUs for 2 epochs. Besides, the loss balancing factor $\lambda$ is set to 1.0. For the inference stage, we attempt to adopt CoT to exploit the advantages of our module design and improve the consistency of reasoning.

% and simplify the P-LoRA to handle each question type, given the lack of comprehensive coverage of driving tasks. We adopt the respective default settings for each VLM, including image size, tokenizer, system prompt, and other relevant parameters. 

% For the inference stage, we attempt to adopt CoT to exploit the advantages of our module design and improve the consistency of reasoning. In detail, we only employ it on the core questions in the DriveLM dataset, such as scene description, comprehensive prediction and ego planning, while still applying individual reasoning for those simple perception and motion estimation.

\subsection{Main results}

We first integrate \netName{} with three representative VLMs and compare the performance of original models and \netName{} versions on DriveLM end-to-end reasoning benchmark. The results on the test split are presented in Table~\ref{tab:drivelm}. \netName{} has improved all these approaches to some extent, only with a lightweight language model and parameter-efficient fine-tuning. Concretely, our method significantly enhances the weaker models like LLaMA-Adapter-V2, particularly showing performance enhancements of 3.44\% on accuracy and 3.89\% on GPT score, respectively. For the powerful VLMs like InternVL, it can still provide decent improvements on overall metrics. In addition, we observed that the gains on match metric of InternVL are markedly higher than others. This can be attributed to the learned localization capability during web-scale pretraining, which facilitates the utilization of end-to-end driving information. These results conclusively demonstrate the effectiveness and superiority of our \netName{}. Besides, to enable an effective comparison with existing methods, we also provide the results on a set of DriveLM following EM-VLM4AD~\cite{gopalkrishnan2024multi} as shown in Table~\ref{tab:drivelmcmp}. In addition, we provide the official leaderboard in the supplementary materials.

Moreover, we also evaluate \netName{} on the nuScenes-QA dataset in Table~\ref{tab:nuscqa}. Considering that different VLM baselines can lead to significant performance differences, we also adopt the same VLM, namely LLaMA-Adapter, to ensure a fair comparison. It can be observed that \netName{} achieves improvements of 2.57\%  on overall accuracy, with 1.99\% and 3.75\% for H0 and H1 metrics, respectively.

\begin{figure*}[t]
    \centering
    \includegraphics[width=1\textwidth]{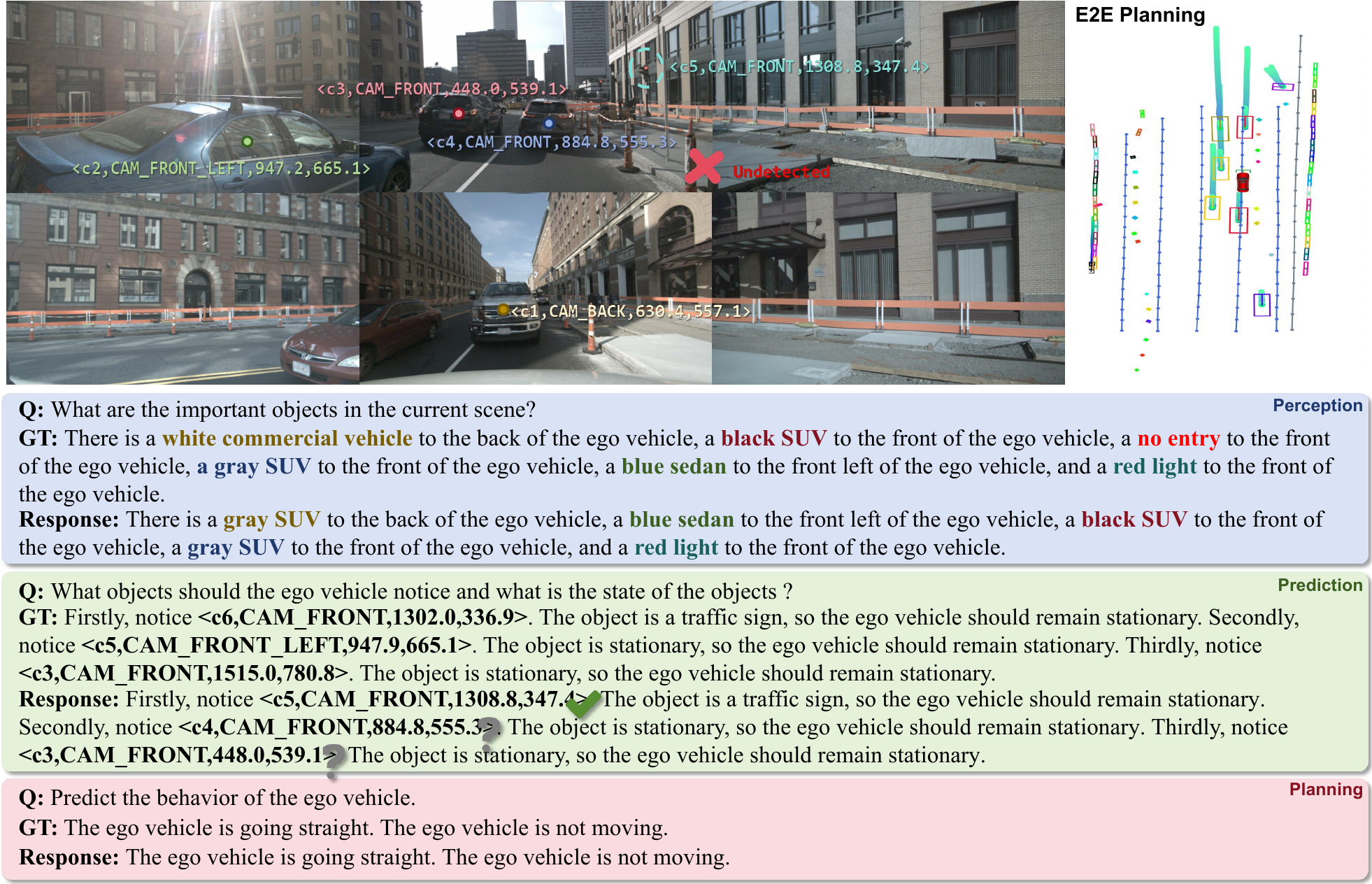}
    \caption{Qualitative results on the {\it DriveLM validation set}. The upper part shows the surrounding multi-view images and the end-to-end planning results on the BEV. Then the three critical questions spanning from perception to planning are provided in the lower part, where question-mark symbol denotes the predictions that, despite being inconsistent with answers, are still reasonable in driving scenarios.}
    \label{fig:vis}
\end{figure*}
\subsection{Ablation study}
We conduct ablation studies on the DriveLM benchmark based on LLaVA-v1.5 model to examine the effectiveness of each component of \netName{}, and perform the ablation on a selected subset of 100 scenes from the DriveLM dataset.

\paragraph{Effects of components.}
As shown in Table~\ref{tab:ab_main}, we assess the effectiveness of each component in our network. The first row (ID-1) represents the baseline VLM, taking multi-view images as input, processing image features with QFormer~\cite{li2023blip2} and fine-tuned with basic LoRA. First, we equip VLMs with the PI encoder to facilitate interaction between surrounding images, leading to a noticeable improvement across all metrics except for the match, as shown in the second row. This could be due to the confusion regarding precise positioning, which hampers the localization capability of VLMs. Then, we introduce P-LoRA fine-tuning, which brings significant performance gains over ID-2. Subsequently, we also integrate end-to-end tokens with PI encoder to analyze the validness. It can be observed that end-to-end prior knowledge indeed advances the simple perception and motion estimation (indicated as accuracy) and match, represented as ID-4. 

% Aggregating all these modules, we achieve the best validation performance as shown in the final row. 

\paragraph{Effects of Parallel LoRA fine-tuning.}
Parallel LoRA fine-tuning can significantly promote the VLMs specialize in specific tasks. Hence, we evaluate the impact of different types of P-LoRA design including task-based, question-based and hierarchical mode in Table~\ref{tab:ab_lora}. The task-based mode is the strategy we mainly employ, which comprises perception, prediction and planning. This strategy can effectively contribute to improvements for all metrics. As for the question-based mode, we categorize LoRA branches more specifically based on the problem type, consuming more GPU memory resources. It brings an unbalanced gain, especially promoting the accuracy and GPT score, which we deem to be caused by severe data imbalance. Hence, we also strive to combine these two modes to provide more detailed knowledge propagation, and it exhibits optimal performance. Nevertheless, the question types differ considerably across different dataset, and it is troublesome to maintain the data balance. Therefore, task-based strategy is the versatile and effective design in our experiments.

\paragraph{Effects of end-to-end tokens.}
In Table~\ref{tab:ab_e2etoken}, we investigate the performance variations with respect to the end-to-end tokens that are fed into the VLMs. We first only deliver the perception tokens to the VLMs, and the results show substantial improvements, where the language score and match metric are almost on par with the best performance. That is to say, perception tasks play a key role in behavior interpretation. Then, we introduce the prediction tokens as indicated in the 3rd row. The performance gains are limited except accuracy, which involves several motion status judgment questions. It might be caused by that 3D position changes can be hardly captured due to lacking of spatial awareness. After providing the ego planning token, VLMs learn the explicit interaction relationships between ego vehicle and the surroundings, deriving more suitable motion behavior.

\subsection{Qualitative analysis}
In Figure~\ref{fig:vis}, we present qualitative results of our network on the DriveLM validation set. We provide the multi-view images and the BEV planning results in the upper part of the figure. Then, we select three representative questions utilized to comprehensively describe the driving scenes and provide reasonable behavior explanation, encompassing perception, prediction and planning task. As for the perception task, we highlight the recognized objects with identical colors in the ground truth text and the response, and the object identities are marked in the images. Due to the relatively precise prior position contained in planning results, the surrounding objects can be easily detected. However, VLMs still have difficulties in addressing special signs such as ``no entry'' in the figure, which are not distinct enough. In the prediction task, it can be seen that VLMs pay more attention to the crucial and influential objects, such as the traffic sign. Although other predictions differ the ground truth objects, they indeed have a significant impact on the ego behavior. Hence, the method is able to aggregate the history context and end-to-end results, finally giving the correct planning behaviors appropriate for the current driving environment.

% More qualitative results and failure cases are available in the supplementary materials.

\vspace{0.5em}

\section{Conclusion}
\label{sec:concl}
% \vspace{0.5em}
In this work, we revisit the implementation of VLMs in the field of autonomous driving and introduce \netName{}, specifically tailored for this domain. The key components of our framework are the preliminary interaction mechanism and Parallel LoRA, which improve scene interaction and understanding, and enable task-specific specialization, respectively. In addition, we integrate end-to-end driving frameworks with \netName{}, providing rich spatial and motion prior information for VLMs. Our extensive experiments, conducted on both DriveLM and nuScenes-QA benchmarks, demonstrate that \netName{} consistently improves several VLM baselines.
% This highlights its potential as a promising approach for explainable driving scenarios in the rapidly evolving field of autonomous driving.

\vspace{0.5em}
\paragraph{Limitations and future work.}
We focus on employing \netName{} for the reasoning task in explainable autonomous driving. However, we believe that this end-to-end paradigm of VLMs is also valuable for planning tasks, and warrants further exploration in future research.
Additionally, achieving complementarity between end-to-end frameworks from the vehicle perspective and VLMs from the human perspective could benefit both sides.

% We anticipate to address this to maximize the benefits of our framework.

%%%%%%%%%%%%%%%%%%%%%%%%%%%%%%%%%%%%%%%%%%%%%%%%%

\bibliography{aaai2026}

%%%%%%%%%%%%%%%%%%%%%%%%%%%%%%%%%%%%%%%%%%%%%%%%%
% \input{sec/7_checklist}
%%%%%%%%%%%%%%%%%%%%%%%%%%%%%%%%%%%%%%%%%%%%%%%%%

%%%%%%%%%%%%%%%%%%%%%%%%%%%%%%%%%%%%%%%%%%%%%%%%%
% \clearpage
\setcounter{page}{1}
% \maketitlesupplementary

% \startcontents
% {
%     \hypersetup{linkcolor=black}
%     \printcontents{}{1}{}
% }
% \newpage
\section*{Appendix}

\section{More implementation details}
% \paragraph{Metrics.}
% We follow the metrics proposed in DriveLM for evaluation, including accuracy, ChatGPT score, language score and the match. The final score is the weighted average of these scores with a weight of 0.2, 0.4, 0.2 and 0.2 respectively. As for the nuScenes-QA dataset, we adopt three metrics evaluated on overall set, H0 (zero-hop) set and H1 (one-hop) set following the official settings. The ChatGPT Score is employed to precisely measure the semantic alignment between ground truth and predicted answers. The Language Score including $\rm BLEU$, $\rm ROUGE\_L$, $\rm CIDEr$ is used for the evaluation of language generation. The Match score evaluates the matched important objects over the all objects.

\paragraph{Details for the utilization of end-to-end tokens.}
With respect to different VLMs, our approach to integrating VLM differs slightly. Specifically, we directly concatenate the end-to-end tokens with vision tokens and form the new adapter for the LLaMA-Adapter model~\cite{gao2023llamaadapterv2}, with the same adapter interaction adopted. As for the LLaVA model~\cite{liu2023llava}, we insert the end-to-end tokens after vision tokens with ``$\backslash {\rm n}$'' as an interval. To integrate with the internVL model~\cite{chen2024internvl}, we introduce additional special tokens ``$<{\rm e2e}>$'' and ``$</{\rm e2e}>$'' as the indicator for end-to-end tokens following vision tokens similarly.

\paragraph{Details for the inference.}
During the inference stage, we also adopt CoT technique to enhance the coherent following~\cite{sima2023drivelm}. However, most questions involve object identities, which provide the ground truth information for subsequent reasoning tasks. To prevent this possible information leakage derived from the questions, we only perform CoT for core questions. Regarding the inference latency and model overhead, the modules introduced account for only 3$\sim$4\% of the total parameters among LMAD variants. That is, the latency and overhead are mostly from the VLM adopted.

\begin{table*} [ht!] 
    \centering
    {\begin{tabular}{c|l}
       \toprule[1.5pt]
        \textbf{Notation} & \textbf{Description} \\\midrule

        $N_q$ & the number of learnable queries of QFormer \\
        $N_c$ & the number of camera queries \\
        $N_{\rm ins}$ & the number of selected instances \\
        $F_{\rm det}$, $F_{\rm mot}$ and $F_{\rm ego}$ & detection, motion and ego features from e2e framework  \\
        $T_{\rm mot}$ and $T_{\rm ego}$ & object prediction trajectories and ego planning trajectories \\
        $F_{\rm det}^{\rm n}$, $F_{\rm mot}^{\rm n}$ and $F_{\rm ego}^{\rm n}$  & numerical features for detection, motion and ego \\
        $F_{\rm det}^{\rm t}$, $F_{\rm mot}^{\rm t}$ and $F_{\rm ego}^{\rm t}$  & textual features for detection, motion and ego \\
        $F_{\rm e2e}$ & the e2e features input into VLMs \\

        \bottomrule[1.5pt]
    \end{tabular}}
    % \vspace{1.5pt}
    \caption{Notations used in the paper.}
    \label{tab:supp_notation}
\end{table*}
\begin{table} [ht!] 
    \centering
    % \resizebox{\columnwidth}{!}{
    \begin{tabular}{cc|cccc}
       \bottomrule[1.5pt]
           num. & txt. & Acc. & GPT & Lan. & Match\\\hline
         & & 72.76 & 62.71 & 45.99 & 34.73\\
         \checkmark & & 73.14 & \bf62.98 & 46.12& 34.77\\
        \rowcolor{gray!20}
         \checkmark & \checkmark & \bf 73.81 & 62.93 & \bf 46.76& \bf 34.85\\
        \bottomrule[1.5pt]
    \end{tabular}
    % }
    % \vspace{1.5pt}
    \caption{Ablation study on the numerical and textual prompt features. ``num." and ``txt." represent the numerical and textual prompt, respectively.}
    \label{tab:ab_numtxt}
\end{table}

\begin{table} [ht!] 
    \centering
        \centering
        % \resizebox{0.98\columnwidth}{!}
        {\begin{tabular}[b]{l|ccc|ccc}
        \toprule[1.5pt]

        \multirow{2}{*}{Method} &
        \multicolumn{3}{c|}{L2 ($m$) $\downarrow$} & 
        \multicolumn{3}{c}{Col. Rate (\%) $\downarrow$} \\
        & 1$s$ & 2$s$ & 3$s$ & 1$s$ & 2$s$ & 3$s$\\
        \midrule 
         VAD-tiny & 0.46 & 0.76 & 1.12 & 0.21 & 0.35 & 0.58 \\
         \rowcolor{gray!20}
         + \netName{}& 0.42 & 0.71 &  1.07 & 0.15 &  0.26 &  0.48 \\
         VAD-base & 0.41 & 0.70 & 1.05  & 0.05 & 0.17 & 0.43  \\
         \rowcolor{gray!20}
         + \netName{} &  0.38 & 0.65 & 0.99 & 0.03 &  0.11 &  0.35 \\
         
         \bottomrule[1.5pt]
        \end{tabular}}
    
    % \vspace{1.5pt}
    \caption{Planning results on the nuScenes validation dataset.}
    \label{tab:e2e}
\end{table}

\paragraph{Notations.}
As shown in Table~\ref{tab:supp_notation}, we provide a lookup table for notations used in the paper.

\section{More experiments}
\subsection{Metrics}
The DriveLM dataset proposes four kinds of metrics for different question types, including accuracy, GPT score, language score and match. The accuracy is the ratio of correctly predicted samples of choice and judgment questions to the total number of samples. The ChatGPT Score is employed to precisely measure the semantic alignment between ground truth and predicted answers, which is mainly utilized on the planning questions. The language score is used for the evaluation of language generation for core perception questions, calculated as the average of $\rm BLEU$, $\rm ROUGE\_L$ and $\rm CIDEr$ scores. The match evaluates the matched important objects over the all objects for core prediction questions, which is composed of matching accuracy and GPT score. The final score is the weighted average of these scores with a weight of 0.2, 0.4, 0.2 and 0.2 respectively. 

As for the nuScenes-QA dataset, the metrics are H0, H1 and overall metrics, all representing the accuracy of correct answers of corresponding questions. Concretely, the questions of H1 (one-hop) group involve reasoning about relations between objects, while H0 (zero-hop) questions are relatively simpler.

\subsection{Leaderboard on DriveLM}
As shown in Table~\ref{tab:supp_drivelm}, we compare our work with others on the DriveLM leaderboard.

\subsection{Adapting \netName{} to a VLA model}
To further demonstrate the generalization performance of \netName{}, we adapt \netName{} to a VLA model, performing planning task instead of reasoning task. Specifically, we intergrade \netName{} with a planning head and remove the original end-to-end branch. The performance on Bench2Drive~\cite{jia2024bench2drive} dataset are demonstrated in Table~\ref{tab:bench}

\subsection{Ablation study}
\paragraph{Effects of numerical and textual prompt.}
In Table~\ref{tab:ab_numtxt}, we assess the influences of the numerical and textual prompts utilized for end-to-end tokens. Compared to the numerical prompts, textual prompt can lead to a more comprehensive improvements, which demonstrates the textual representations are more comprehensible for VLMs. 

\paragraph{Joint training with end-to-end driving}
\label{exp:joint}

Since end-to-end features have demonstrated the effect on VLMs, we expect it to be effective in reverse as well. To fulfill this, we first adjust the DriveLM dataset to make it better aligned with end-to-end driving tasks. Specifically, we first reorganize the dataset by selecting the questions about comprehensive scene description and removing others less relevant to end-to-end planning. Besides, DriveLM only involves text annotations for partial crucial scenes, which are insufficient for training. To address this limitations, we expand the QA datasets and introduce extra labels for the neighbors of these scenes by adjusting the position in object identities.

Subsequently, we activate the gradient flow from VLMs to end-to-end driving framework, and consider VLMs as an auxiliary module for supervision. Due to the memory constraint and module adaptability, we only adopt VAD~\cite{jiang2023vad} as the baseline in this work, and fine-tune the task heads with pretrained weights for simplicity. As shown in Table~\ref{tab:e2e}, the improvements for the L2 error are marginal, possibly caused by the lack of numerical motion and planning QAs in the reasoning dataset. Thereby, it struggles to meet the requirements for the precise trajectory planning. In contrast, decent gains are observed on the collision rate, which we attribute to the supplementary scene awareness involved in the text and the novel human perspective.

\begin{figure}
    \centering
    \includegraphics[width=0.47\textwidth]{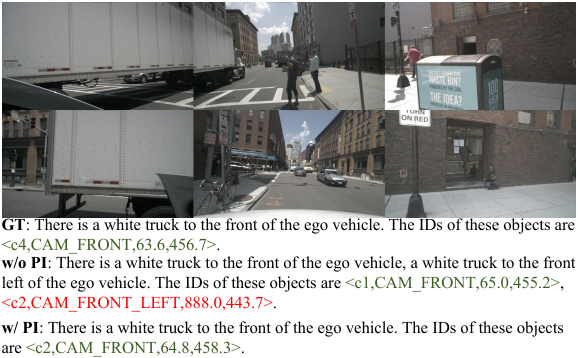}
    \caption{Quantitative improvements of our PI module.}
    \label{fig:crossview}
\end{figure}

\begin{table*} [ht!] 
    \centering
        \centering
        \resizebox{0.98\textwidth}{!}
        {\begin{tabular}[b]{l|ccccccccc|c}
        \toprule[1.5pt]

        \textbf{Method} & Acc. & GPT & $\rm BLEU_1$ & $\rm BLEU_2$ &$\rm BLEU_3$ &$\rm BLEU_4$ &$\rm ROUGE\_L$ & $\rm CIDEr$ & Match & Final \\

        \midrule

        DriveMM~\cite{huang2024drivemm} & 76.09 & 66.44 & 0.7844& 0.7232 & 0.6664 & 0.6115 & 0.7477 & 0.2261 & 48.63 & 61.30 \\

        LMAD & 80.38 & 65.10 & 0.7867& 0.7232 & 0.6640 & 0.6066 & 0.7410 & 0.1781 & 46.12 & 61.03 \\

        LDM & 76.50 & 65.57 & 0.7335 & 0.6751 & 0.6206 & 0.5677 & 0.7297 & 0.169 & 49.20 & 60.68 \\
        
        E2EAutoMatrixTeam & 76.25 & 64.11 & 0.7633 & 0.7011 & 0.6425 & 0.586 & 0.7405 & 0.2081 & 50.89 & 60.64 \\
        
        DrivingLanguage & 75.92 & 65.18 & 0.6987 & 0.649 & 0.5997 & 0.5497 & 0.7146 & 0.3900 & 48.34 & 60.11 \\

        DrivingOV & 74.73 & 64.98 & 0.7623 & 0.7015 & 0.6446 & 0.5891 & 0.7418 & 0.1806 & 44.94 & 59.49 \\

        hdhdhd & 81.48 & 65.45 & 0.7903 & 0.7281 & 0.6696 & 0.6128 & 0.7476 & 0.1982 & 33.86 & 59.03 \\

        MTSU & 75.20 & 60.97 & 0.7802 & 0.7181 & 0.6603 & 0.6048 & 0.7405 & 0.1972 & 48.83 & 58.87 \\

         \bottomrule[1.5pt]
        \end{tabular}}
    
    % \vspace{1.5pt}
    \caption{DriveLM official leaderboard.}
    \label{tab:supp_drivelm}
\end{table*}
\begin{table} [t!] 
    \centering
    {\begin{tabular}{l|cc}
       \bottomrule[1.5pt]
        $\textbf{Method}$  & DS & SR(\%)\\\hline
        VAD~\cite{jiang2023vad}&
        42.35 &15.00 \\ 
        GenAD~\cite{zheng2024genad}&
        44.81 & 15.90\\
        MomAD~\cite{song2025momad}&
        44.54 & 16.71\\
        DriveTransformer~\cite{jia2025drivetransformer}&
        63.46 & 35.01\\
        Orion~\cite{fu2025orion}&
        77.74 &  54.62\\

        \Xhline{0.9pt} 
        \netName{}       & \bf 80.18 & \bf 59.60  \\\bottomrule[1.5pt]
    \end{tabular}}
    % \vspace{1.5pt}
    \caption{Closed-loop performance comparison on Bench2Drive dateset. DS: Driving Score, SR: Success Rate}
    \label{tab:bench}
\end{table}

\begin{figure*}[t]
    \centering
    \includegraphics[width=0.98\textwidth]{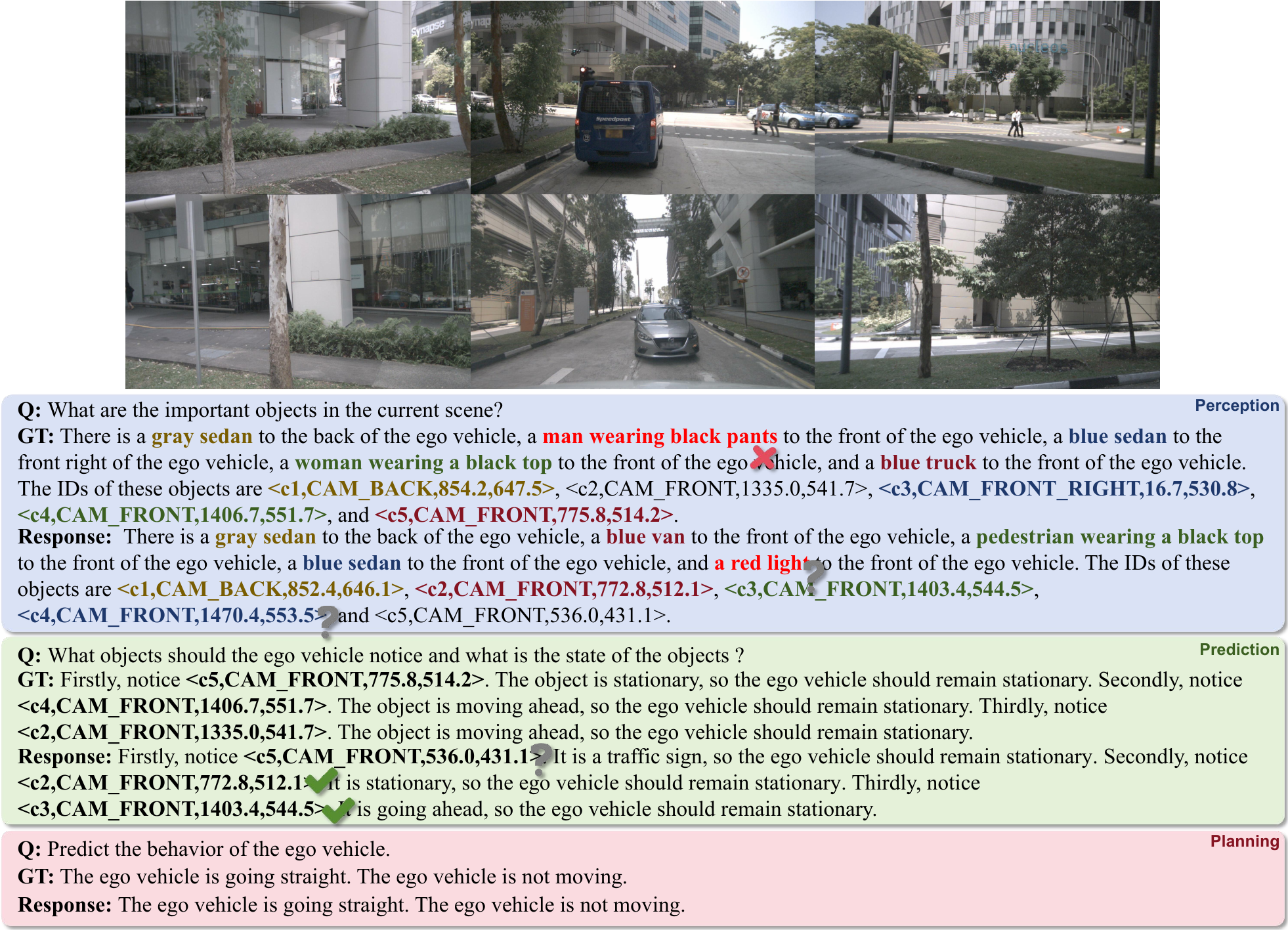}
    \caption{Qualitative results on the {\it DriveLM validation set}.}
    \label{fig:vis_sup}
\end{figure*}
\begin{figure*}[t]
    \centering
    \includegraphics[width=0.98\textwidth]{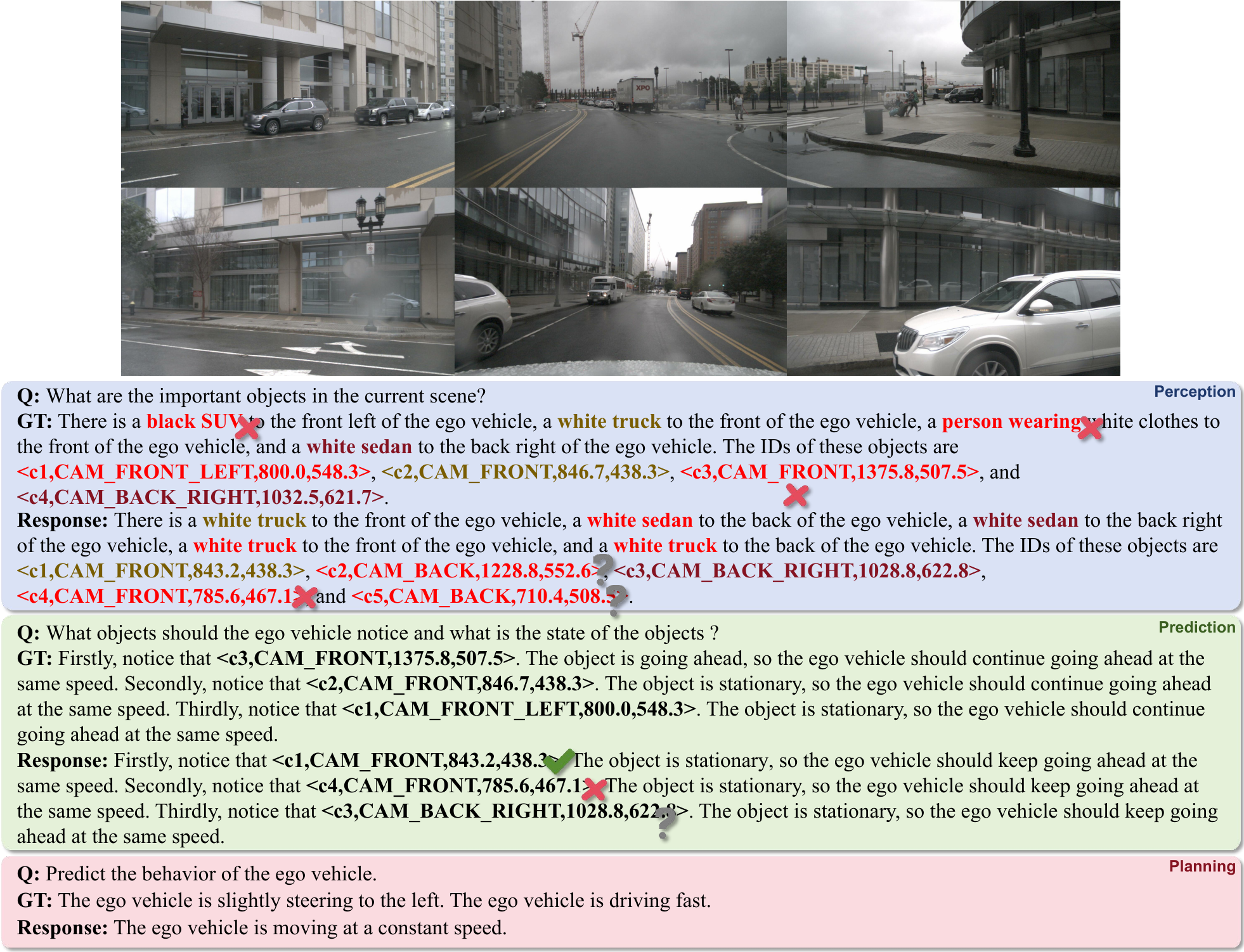}
    \caption{Failure cases on the {\it DriveLM validation set}.}
    \label{fig:vis_sup_fail}
\end{figure*}
\section{More qualitative results}

\paragraph{Quantitative improvements of PI module.}
The PI module can be integrated with both QFormer-based and direct input-based methods. To further verify the benefits of PI module, we also provide a qualitative result in the Figure~\ref{fig:crossview}, showing that scene-level interaction helps alleviate ambiguous cases (cross-image objects).

\section{More qualitative results of \netName{}}
We present additional qualitative results in Figure~\ref{fig:vis_sup} and Figure~\ref{fig:vis_sup_fail}. As shown in Figure~\ref{fig:vis_sup}, our framework successfully detects most key objects, with the exceptions of ``a man wearing black pants'' and ``a blue sedan''. Notably, the blue sedan appears in two images, and \netName{} is able to accurately identify and select the more relevant part through scene PI. Moreover, it also detects the traffic sign, which plays a more significant role in influencing ego vehicle behavior but is not included in the ground truth. The failure cases are depicted in Figure~\ref{fig:vis_sup_fail}, particularly under rainy conditions. In this case, \netName{} may produce incorrect detections, such as ``c4'', or identify low-impact objects like ``c2'' and ``c5'', which can further hinder subsequent prediction and planning tasks.
%%%%%%%%%%%%%%%%%%%%%%%%%%%%%%%%%%%%%%%%%%%%%%%%%

\end{document}